\documentclass{article}
\pdfoutput=1
\usepackage[top=1.2in, left=1.4in, right=1.4in, includefoot]{geometry}
\usepackage{fancyhdr}
\usepackage{graphicx}
\usepackage{times}
\usepackage{xcolor}
\usepackage{epsfig}
\usepackage{caption}
\usepackage{subcaption}
\usepackage{graphicx}
\usepackage{booktabs}
\usepackage{amsmath, amssymb, amsfonts}
\usepackage{mathtools, mathrsfs}
\usepackage[ruled, vlined]{algorithm2e}
\usepackage{bm}
\usepackage[mathscr]{eucal}
\usepackage[pagebackref=true,breaklinks=true,colorlinks,bookmarks=false]{hyperref}
\hypersetup{
    colorlinks,
    linkcolor={red!75!black},
    citecolor={blue!80!black},
    urlcolor={blue!80!black}
}
\usepackage[round]{natbib}

\newcommand\mX{\mathscr{X}}

\renewcommand{\b}{\mathbf}
\newcommand{\tb}{\textbf}

\newcommand\R{\mathbb{R}}
\newcommand\mY{\mathscr{Y}}

\newcommand\mH{\mathscr{H}}

\newcommand\mR{\mathscr{R}}
\newcommand\mS{\mathscr{S}}

\newcommand\mL{\mathscr{L}}

\newcommand\risk{\mathscr{R}}
\newcommand\riskemp{\risk_\mS}
\newcommand\reals{\mathbb{R}}

\newcommand{\T}{\top}         
\newcommand{\N}{\mathbb{N}}   
\DeclareMathOperator{\Tr}{Tr} 

\newtheorem{theorem}{Theorem}[section]
\newtheorem{lemma}[theorem]{Lemma}

\newtheorem{proof}[theorem]{Proof}

\begin{document}

\title{Emotion Transfer Using Vector-Valued Infinite Task Learning}

\author{Alex Lambert$^{1,}$\thanks{Both authors contributed equally. $1$: LTCI, T{\'e}l{\'e}com Paris, Institut Polytechnique de Paris, France. $2$: Center of Applied Mathematics, CNRS, École Polytechnique, Institut Polytechnique de Paris, France. Corresponding author: alex.lambert@telecom-paris.fr} \and Sanjeel Parekh$^{1,*}$ \and Zolt{\'a}n Szab{\'o}$^{2}$ \and Florence d'Alch\'e-Buc$^{1}$}

\date{}

\maketitle

\abstract{
 Style transfer is a significant problem of machine learning with numerous successful applications. In this work, we present a novel style transfer framework building upon infinite task learning and  vector-valued reproducing kernel Hilbert spaces. We instantiate the idea in emotion transfer where the goal is to transform  facial images to different target emotions. The proposed approach provides a principled way to gain explicit control over the continuous style space. We demonstrate the efficiency of the technique on popular facial emotion benchmarks, achieving low reconstruction cost and high emotion classification accuracy.
}

\section{Introduction}
\label{section:introduction}

Recent years have witnessed an increasing attention around style transfer problems \citep{gatys2016,wynen18unsupervised,jing2020-review} in machine learning. In a nutshell, style transfer refers to the transformation of an object according to a target style. It has found numerous applications in computer vision \citep{ulyanov2016style,choi2018stargan,puy19flexible,yao20high}, natural language processing \citep{fu2018} as well as audio signal processing \citep{grinstein2018} where objects at hand are contents in which style is inherently part of their perception. Style transfer is one of the key components of data augmentation \citep{mikolajczyk2018} as a means to artificially generate meaningful additional data for the training of deep neural networks. Besides, it has also been shown to be useful for counterbalancing bias in data by producing stylized contents with a well-chosen style (see for instance \cite{geirhos2019}) in image recognition. More broadly,  style transfer fits into the wide paradigm of parametric modeling, where a system, a process or a signal can be controlled by its parameter value. Adopting this perspective, style transfer-like applications can also be found in digital twinning \citep{tao19digital,rita19survey,lim20state}, a field of growing interest in health and industry.

In this work, we propose a novel principled approach for style transfer, exemplified in the context of emotion transfer of face images. Given a set of emotions, classical emotion transfer refers to the task of transforming face images according to these target emotions. The pioneering works in emotion transfer include that of \citet{blanz1999morphable} who proposed a morphable 3D face model whose parameters could be modified for facial attribute editing. \citet{susskind2008generating} designed a deep belief net for facial expression generation using action unit (AU) annotations.

More recently, extensions of generative adversarial networks (GANs, \citealt{goodfellow2014generative}) have proven to be particularly powerful for tackling image-to-image translation problems \citep{zhu2017unpaired}. Several works have addressed emotion transfer for facial images by conditioning GANs on a variety of guiding information ranging from discrete emotion labels to photos and videos. In particular, StarGAN \citep{choi2018stargan} is conditioned on discrete expression labels for face synthesis. ExprGAN \citep{ding2018exprgan} proposes synthesis with the ability to control expression intensity through a controller module conditioned on discrete labels. Other GAN-based approaches make use of additional information such as AU labels \citep{pumarola2018ganimation}, target landmarks \citep{ qiao2018geometry}, fiducial points \citep{ song2018geometry} and photos/videos \citep{geng2018warp}.
While GANs have achieved high quality image synthesis, they come with some pitfalls: they are particularly difficult to train and require large amounts of training data. 

In this paper, unlike previous approaches, we adopt a functional point of view: given some person, we assume that the full range of the emotional faces can be modelled as a continuous function from emotions to images. This view exploits the geometry of the representation of emotions \citep{russell1980circumplex}, assuming that one can pass  a facial image ``continuously" from one emotion to an other.
We then propose to address the problem of emotion transfer by learning an image-to-function model able to predict for a given facial input image represented by its landmarks \citep{tautkute2018}, the continuous function that maps an emotion  to the image transformed by this emotion. 

This function-valued regression approach relies on a technique recently introduced by \citet{brault2019infinite} called infinite task learning (ITL). ITL enlarges the scope of multi-task learning \citep{evgeniou04regularized,evgeniou05learning} by learning to solve simultaneously a set of tasks parametrized by a continuous parameter. While strongly linked to other parametric learning methods such the one proposed by \citet{takeuchi06nonparametric}, the approach differs from previous works by leveraging the use of operator-valued kernels and vector-valued reproducing kernel Hilbert spaces (vRKHS; \citealt{pedrick1957theory, micchelli05learning, carmeli06vector}). vRKHSs have proven to be relevant in solving supervised learning tasks such as multiple quantile regression \citep{sangnier16joint} or unsupervised problems like anomaly detection \citep{scholkopf2001estimating}. A common property of these works is that the output to be predicted is a real-valued function of a real parameter.

To solve the emotion transfer problem, we present an extension of ITL, vector ITL (or shortly vITL) which involves functional outputs with vectorial representation of the faces and the emotions, showing that the approach is still easily controllable by the choice of appropriate kernels guaranteeing continuity and smoothness. In particular, the functional point of view by the inherent regularization induced by the kernel makes the approach suitable even for limited and partially observed emotional images.  We demonstrate the efficiency of the vITL approach in a series of numerical experiments showing that it can achieve state-of-the-art performance on two benchmark datasets.

The paper is structured as follows. We formulate the problem and introduce the vITL framework in Section~2. Section~3 is dedicated to the underlying optimization problem. Numerical experiments conducted on two benchmarks of the domain are presented in Section~4. Discussion and future work conclude the paper in Section~5. Proofs of auxiliary lemmas are collected in Section~6.

\section{Problem Formulation}
\label{section:problem_setting}

In this section we define our problem. Our aim is to design a system capable of transferring emotions: having access to the face image of a given person our goal is to convert his/her face to a specified target emotion. In other words, the system should implement a mapping of the form
\begin{align}
    \text{(face, emotion)} \mapsto \text{ face}.
    \label{eq:face,emotion->face}
\end{align}
In order to tackle this task, one requires a representation of the emotions, and similarly that of the faces. The classical categorical description of emotions deals with the classes `happy', `sad', `angry', `surprised', `disgusted', `fearful'. The valence-arousal model \citep{russell1980circumplex} embeds these categories into the $2$-dimensional space. The resulting representation of the emotions are points $\theta \in \R^2$, each coordinate of these vectors encoding the valence (pleasure to displeasure) and arousal (high to low) associated to the emotions. This is the emotion representation we use while noting that there are alternative encodings in higher dimension ($\Theta \subset \R^p$, $p\ge 2$; \citealt{vemulapalli2019compact}) to which the presented framework can be naturally adapted. Throughout this work faces are represented by landmark points. Landmarks have been proved to be a useful representation in facial recognition \citep{saragih2009, scherhag2018detecting, zhang2015pose}, 3D facial reconstruction and sentiment analysis. \citet{tautkute2018} have shown that emotions can be accurately recognized by detecting changes in the localization of the landmarks. Given $M$ number of landmarks on the face, this means a description $x \in \mX:=\R^{2M=:d}$.
\begin{figure}
	\centering
	\includegraphics[scale=0.6]{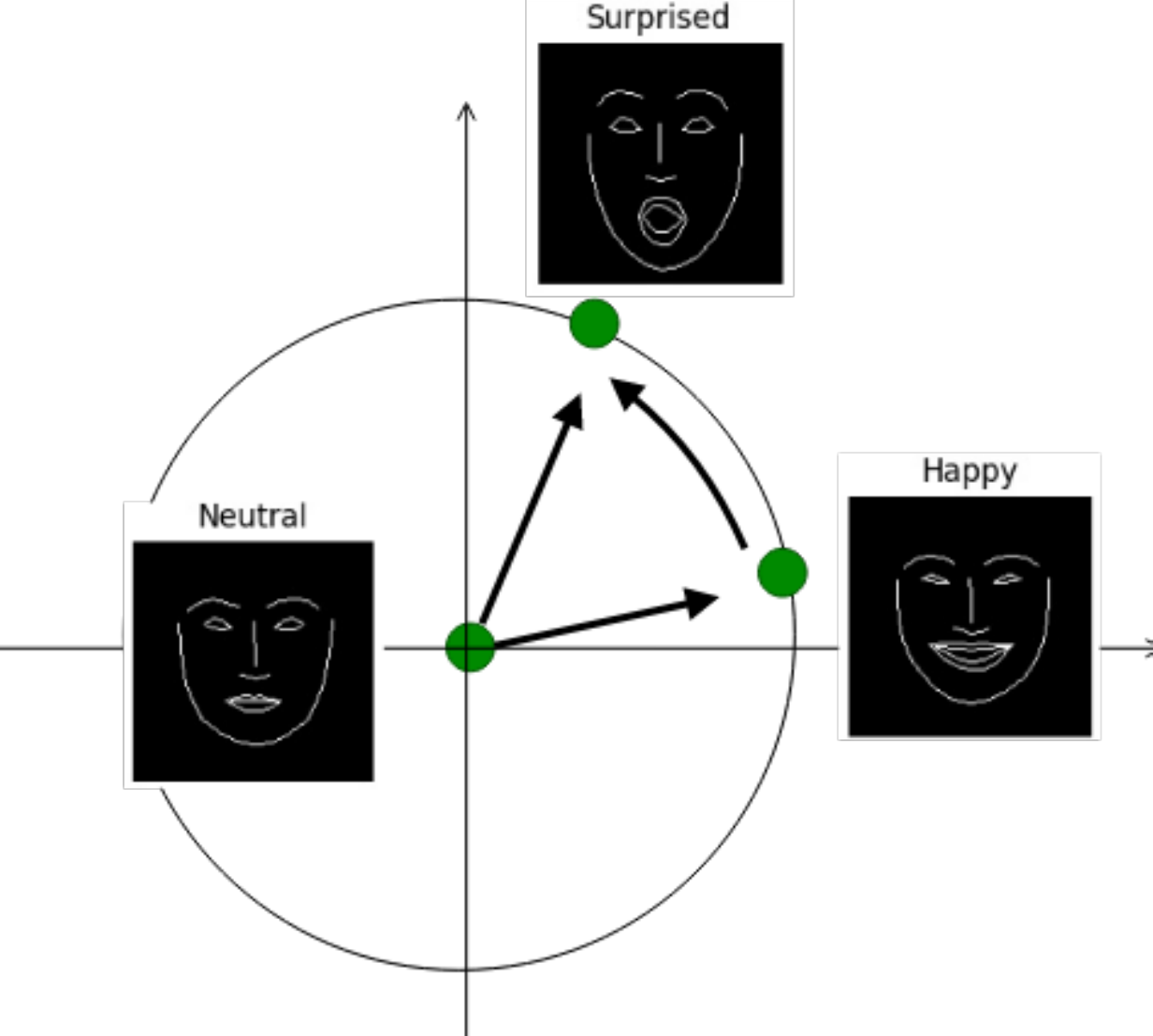}
	\caption{Illustration of emotion transfer.}
	\label{va-space}
\end{figure}
The resulting mapping \eqref{eq:face,emotion->face} is illustrated in Fig.~\ref{va-space}: starting from a neutral face and the target happy one can traverse to the happy face; from the happy face, given the target emotion surprise one can get to the surprised face.

In an ideal world, for each person, one would have access to a trajectory $z$ mapping each emotion $\theta \in \Theta$ to the corresponding landmark locations $ x \in \mX$; this function $z:\Theta \mapsto \mX$ can be taken for instance to be the element of $L^2[\Theta, \mu; \mX]$, the space of $\R^d$-valued square-integrable function w.r.t.\ to a measure $\mu$.
The probability measure $\mu$ allows capturing the frequency of the individual emotions. In practice, one has realizations $(z_i)_{i\in[n]}$, each $z_i$ corresponds to a single person possible appearing multiple times. The trajectories  are observable at finite many emotions $\left (\tilde{\theta}_{i,j}\right)_{j\in [m]}$ where $[m]:=\{1,\ldots,m\}$.\footnote{To keep the notation simple, we assume that $m$ is the same for all the $z_i$-s.} In order to capture relation \eqref{eq:face,emotion->face} one can rely on a hypothesis space $\mH$ with elements
\begin{align}
     h:  \mX \mapsto (\Theta \mapsto \mX). \label{eq:h}
\end{align}
The value $ h(x)(\theta)$ represents the landmark prediction from face $x$ and target emotion $\theta$.

We consider \tb{two tasks} for emotion transfer:
\begin{itemize}
    \item \tb{Single emotional input:} In the first problem, the assumption is that all the faces appearing as the input in \eqref{eq:face,emotion->face} come from a fixed emotion $\tilde{\theta}_0 \in \Theta$. The data which can be used to learn the mapping $h$ consists of $t=n$ triplets\footnote{In this case $\theta_{i,j}$ is a literal copy of $\tilde{\theta}_{i,j}$ which helps to get a unified formulation with the joint emotional input setting.}
     \begin{align*}
       x_i &= z_i \big(\tilde{\theta}_0 \big) \in \mX, & \b Y_i & = \big(\underbrace{z_i \big (\tilde{\theta}_{i,j} \big)}_{=:y_{i,j}}\big)_{j\in [m]}\in \mX^m, \\ (\theta_{i,j})_{j \in [m]} &= \big(\tilde{\theta}_{i,j}\big)_{j \in [m]} \in \Theta^m,\, i \in [t].
     \end{align*}
     To measure the quality of the reconstruction using a function $h$, one can consider a convex loss $\ell \colon \mX \times \mX \to \R_+$ on the landmark space where $\R_+$ denotes the set of non-negative reals. The resulting objective function to minimize is
     \begin{equation} \label{eq:riskemp}
         \riskemp(h) := \frac{1}{tm} \sum_{i \in [t]} \sum_{j \in [m]} \ell(h(x_i)(\theta_{i,j}),y_{i,j}).
     \end{equation}
     The risk $\riskemp(h)$ captures how well the function $h$ reconstructs on average the landmarks $y_{i,j}$ when applied to the input landmark locations $x_i$.
    \item \tb{Joint emotional input:} In this problem, the faces appearing as input in \eqref{eq:face,emotion->face} can arise from any emotion. The observations consist of triplets
     \begin{align*}
         x_{m(i-1)+l} &= z_{i}\big (\tilde{\theta}_{i,l} \big) \in \mX, & \b Y_{m(i-1)+l} &= \big(\hspace{-0.25cm}\underbrace{z_i\big(\tilde{\theta}_{i,j} \big)}_{=:y_{m(i-1)+l,j}}\hspace{-0.25cm}\big)_{j\in [m]}  \in \mX^m\\
         (\theta_{m(i-1)+l,j})_{j \in [m]} &= \big (\tilde{\theta}_{i,j} \big )_{j \in [m]} \in \Theta^m,
     \end{align*}
     where $(i,l) \in [n]\times [m]$ and the number of pairs  is $t=nm$.
     Having defined this dataset one can optimize the same objective \eqref{eq:riskemp} as before. Particularly, this means that the pair $(i,l)$ plays the role of index $i$ of the previous case. The $(\theta_{i,j})_{i,j \in [t]\times [m]}$ is an extended version of the $\big(\tilde{\theta}_{i,j}\big)_{i,j \in [t]\times [m]}$ to match the indices going from $1$ to $t$ in \eqref{eq:riskemp}.
\end{itemize}

We leverage the flexible class of vector-valued reproducing kernel Hilbert spaces (vRKHS; \citet{carmeli2010vector}) for the hypothesis class schematically illustrated in \eqref{eq:h}. Learning within vRKHS has been shown to be relevant for tackling function-valued regression \citep{Kadri2010,kadri2016ovk}. The construction follows the structure
\begin{align}
     h:  \underbrace{\mX \mapsto \underbrace{(\Theta \mapsto \mX)}_{\in \mH_G}}_{\in \mH_K} \label{eq:h2}
\end{align}
which we detail below. The vector ($\R^d$)-valued capability is beneficial to handle the $\Theta \mapsto \mX=\R^d$ mapping; the associated $\R^d$-valued RKHS $\mH_G$ is uniquely determined by a matrix-valued kernel $G:\Theta\times \Theta \rightarrow \R^{d\times d} = \mL(\mX)$ where $\mL(\mX)$ denotes the space of bounded linear operators on $\mX$, in this case the set of $d\times d$-sized matrices. Similarly, in \eqref{eq:h2} the $\mX \rightarrow \mH_G$ mapping is modelled by a vRKHS $\mH_K$ corresponding to an operator-valued kernel $K: \mX \times \mX \to \mL(\mH_G)$. A matrix-valued kernel ($G$) has to satisfy two conditions: $G(\theta,\theta')=G(\theta',\theta)^\T$ for any $(\theta,\theta')\in \Theta^2$ where $(\cdot)^\T$ denotes transposition, and $\sum_{i,j\in [N]} v_i^\T G(\theta_i,\theta_j) v_j \geq 0$ for all $N\in \N^*:=\{1,2,\ldots\}$, $\{\theta_i\}_{i\in [N]} \subset \Theta$ and $\{v_i\}_{i\in [N]} \subset \R^d$. Analogously, for an operator-valued kernel ($K$) it has to hold that $K(x,x')=K(x',x)^{*}$ for all $(x,x')\in \mX^2$ where $(\cdot)^*$ means the adjoint operator, and $\sum_{i,j\in [N]} \left<w_i, K(x_i,x_j) w_j\right>_{\mH_G} \geq 0$ with  $\left<\cdot,\cdot\right>_{\mH_G}$ being the inner product in $\mH_G$, for all $N\in \N^*$, $\{\theta_i\}_{i\in [N]} \subset \Theta$ and $\{w_i\}_{i\in [N]} \subset \mH_G$ . These abstract requirements can be guaranteed for instance by the choice (made throughout the manuscript)
\begin{align}
    G(\theta,\theta') &= k_{\Theta}(\theta,\theta') \b{A}, &     K(x,x') &= k_{\mX}(x,x') \text{Id}_{\mH_G} \label{eq:G,K:def}
\end{align}
with a scalar-valued kernel $k_{\mX} : \mX \times \mX \to \R$ and $k_{\Theta} : \Theta \times \Theta \to \R$, and symmetric, positive definite matrix $\b A\in \R^{d\times d}$; $\text{Id}_{\mH_G}$ is the identity operator on $\mH_G$.
This choice corresponds to the intuition that for similar input landmarks and target emotions, the predicted output landmarks should also be similar, as measured by $k_{\mX}$, $k_{\Theta}$  and  $\b{A}$, respectively. More precisely, smoothness (analytic property) of the emotion-to-landmark output function can be induced for instance by choosing a Gaussian kernel $ k_{\Theta}(\theta,\theta')= \exp(-\gamma \| \theta - \theta'\|_2^2)$ with $\gamma>0$. The matrix $\b A$ when chosen as $\b A=\b I_d$ corresponds to independent landmarks coordinates while other choices encode prior knowledge about the dependency among the landmarks coordinates \citep{Alvarez2012}. Similarly, the smoothness of function $h$ can be driven by the choice of a Gaussian kernel over $\mX$ while the identity operator on $\mH_G$ is the simplest choice to cope with functional outputs.  By denoting the norm in $\mH_K$ as $\left\|\cdot\right\|_{\mH_K}$, the final objective function is
\begin{align}\label{emp-ivtl}
\min_{h \in \mH_K} \mR_{\lambda}(h) &:= \riskemp(h) + \frac{\lambda}{2} \left\|h\right\|^2_{\mH_K}
\end{align}
with a regularization parameter $\lambda>0$ which balances between the data-fitting term ($\riskemp(h)$) and smoothness ($\left\|h\right\|^2_{\mH_K}$). We refer to \eqref{emp-ivtl} as vector-valued infinite task learning (vITL).

\noindent\tb{Remark:} This problem is a natural adaptation of the ITL framework \citep{brault2019infinite} learning with operator-valued kernels mappings of the form $\mX \mapsto (\Theta \mapsto \mY)$ where $\mY$ is a subset of $\R$; here $\mY = \mX$. An other difference is $\mu$: in ITL this probability measure is designed to approximate integrals via quadrature rule, in vITL it captures the observation mechanism.

\section{Optimization}
This section is dedicated to the solution of  \eqref{emp-ivtl} which is an optimization problem over functions ($h\in \mH_K$). The following  representer lemma provides a finite-dimensional parameterization of the optimal solution. 
\begin{lemma}[Representer]\label{lemma:double-repr}
  Problem \eqref{emp-ivtl} has a unique solution $\hat{h}$ and it takes the form
\begin{equation}
  \hat{h}(x)(\theta) = \sum_{i=1}^t \sum_{j=1}^m k_{\mX}(x,x_i) k_{\Theta}(\theta,\theta_{i,j}) \mathbf{A} \hat{c}_{i,j}, \quad \forall (x,\theta) \in \mX \times \Theta \label{eq:hx-theta}
\end{equation}
for some coefficients $\hat{c}_{i,j}\in \R^d$ with $i\in[t]$ and $j\in [m]$.
\end{lemma}

Based on this lemma finding $\hat{h}$ is equivalent to determining the coefficients $\{\hat{c}_{i,j}\}_{i\in [t],j\in [m]}$. Throughout this paper we consider the squared loss
$\ell (x,x')= \frac{1}{2} \| x-x'\|_2^2$; in this case the task boils down to the solution of a linear equation as detailed in the following result.
\begin{lemma}[optimization task for $\b C$]
  \label{thm:krr}
  Assume that $\b K$ is invertible and let the matrix $\hat{\b C} = [\hat{\b C}_{i}]_{i \in [tm]} \in \R^{(tm) \times d}$ containing all the coefficients, the Gram matrix $\b K = [k_{i,j}]_{i,j \in [tm]} \in \R^{(tm)\times (tm)}$, and the matrix consisting of all the observations $\b Y = [\b{Y}_{i}]_{i \in [tm]} \in \R^{(tm)\times d}$ be defined as 
  \begin{align*} 
    \hat{\b C}_{m(i-1) + j} &:= \hat{c}_{i,j}^\T, \, (i,j) \in [t] \times [m],\\
    k_{m(i_1-1)+j_1, m(i_2 -1)+ j_2} &:= k_{\mX}(x_{i_1},x_{i_2}) k_{\Theta}(\theta_{i_1,j_1},\theta_{i_2,j_2}), \, 
    (i_1,j_1), (i_2, j_2) \in [t] \times [m], \\
    \b Y_{m(i-1) + j} &:= y_{i,j}^\T,\, (i,j) \in [t] \times [m].
\end{align*}
Then $\hat{\b C}$ is the solution of the following linear equation
  \begin{equation} \label{eq:sylvester}
    \b{K}\hat{\b C}\b{A} + tm \lambda \hat{\b{C}} = \b{Y}.
  \end{equation}
 When $\b{A}=\b I_d$ (identity matrix of size $d\times d$), the solution is analytic:
  \begin{equation} \label{eq:ridge_solution}
    \b{\hat{C}} = \left(\b{K} + tm \lambda \b{I}_{tm}\right)^{-1} \b{Y}.
  \end{equation}
\end{lemma}

\noindent \tb{Remarks:}
\begin{itemize}
    \item Computational complexity: In case of $\b A=\b I_d$, 
         the complexity of the closed form solution is $\mathcal{O}\left((tm)^3\right)$. If all the samples are observed at the same locations $(\theta_{i,j})_{i,j \in [t] \times [n]}$, i.e.\ $\theta_{i,j} = \theta_{l,j}$ for $\forall (i,l,j) \in [t] \times [t] \times [m]$, then the Gram matrix $\b{K}$ has a tensorial structure
         $\b{K} = \b K_{\mX} \otimes \b{K}_{\Theta}$ with $\b K_{\mX}=[k_{\mX}(x_i,x_j)]_{i,j\in [t]} \in \R^{t\times t}$ and $\b{K}_{\Theta}=[k_{\Theta}(\theta_{1,i}, \theta_{1,j})]_{i,j\in [m]} \in \R^{m\times m}$. In this case, the computational complexity reduces to $\mathcal{O}\left(t^3 + m^3\right)$. If additional scaling is required one can leverage recent dedicated kernel ridge regression solvers \citep{rudi2017falkon,meanti2020kernel}. 
        If $\b A$ is not identity, then multiplying \eqref{eq:sylvester} with $\b A^{-1}$ gives
        $\b K \hat{\b{C}} + tm \lambda \hat{\b{C}}\b{A}^{-1} = \b{Y} \b{A}^{-1}$ which is a Sylvester equation for which efficient custom solvers exist \citep{el2002block}.
  \item Regularization in vRKHS: 
        Using the notations above, for any $h \in \mH_K$ parameterized by a matrix $\b{C}$, it holds that $ \|h\|^2_{\mH_K} = \Tr\left(\b{K C A C}^\T\right)$. Given  two matrices $\b{A}_1, \b{A}_2$ and associated vRKHSs $\mH_{K_1}$ and $\mH_{K_2}$, if $\b{A}_1$ and $\b{A}_2$ are invertible then any function in $\mH_{K_1}$ parameterized by $\b{C}$ also belongs to $\mH_{K_2}$ (and vice versa), within which it is parameterized by $\b{C} \b{A}_2^{-1} \b{A}_1$. This means that the two spaces contain the same functions, but their norms are different.
\end{itemize}

\section{Numerical Experiments}
\label{section:experiments}
In this section we demonstrate the efficiency of the proposed vITL technique in emotion transfer.
We first introduce the two benchmark datasets we used in our experiments and give details about data representation and choice of the hypothesis space in Section \ref{expt:setup}.
Then, in Section \ref{expt:exp1}, we provide a quantitative performance assessment of the vITL approach (in mean squared error and classification accuracy sense) with a comparison  to the state-of-the-art StarGAN method.
Section \ref{expt:exp2} is dedicated to investigation of the role of $\b A$ (see \eqref{eq:G,K:def})  and the robustness of the approach w.r.t.\ partial observation.
These two sets of experiments (Section~\ref{expt:exp1} and Section~\ref{expt:exp2}) are augmented with a qualitative analysis (Section~\ref{expt:qualitative}). The code written for all these experiments is available on \href{https://www.github.com/allambert/torch_itl}{GitHub}.
\subsection{Experimental Setup}\label{expt:setup}
\begin{figure}
    \begin{subfigure}[b]{\textwidth}
   \includegraphics[scale=0.44]{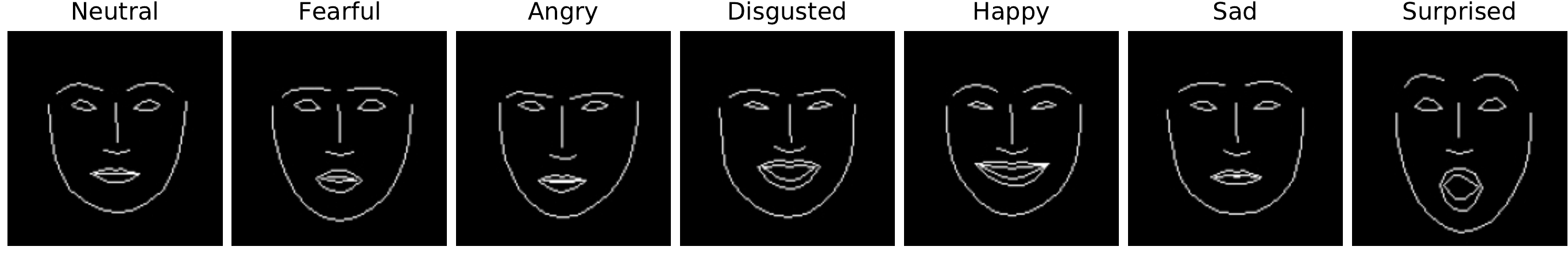}
   \caption{KDEF}
    \end{subfigure}

    \begin{subfigure}[b]{\textwidth}
   \includegraphics[scale=0.44]{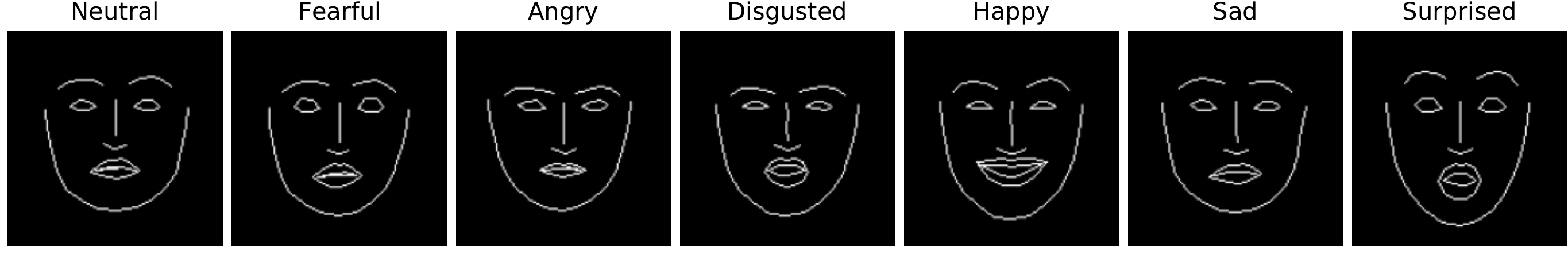}
   \caption{RaFD}
    \end{subfigure}
	\caption{Illustration of the landmark edge maps for different emotions and both datasets.}
	\label{em_illustration}
\end{figure}

We used the following two popular face datasets for evaluation. \begin{itemize}
	\item Karolinska Directed Emotional Faces (KDEF;         \citealt{lundqvist1998karolinska}): This dataset contains facial emotion pictures from $70$ actors ($35$ females and $35$ males) recorded over two sessions which give rise to a total of $140$ samples per emotion. In addition to neutral, the captured facial emotions include afraid, angry, disgusted, happy, sad and surprised.
	\item Radboud Faces Database (RaFD; \citealt{langner2010presentation}): This benchmark contains emotional pictures of $67$ unique identities (including Caucasian males and females, Caucasian children, and Moroccan Dutch males). Each subject was trained to show the following expressions: anger, disgust, fear, happiness, sadness, surprise, contempt, and neutral according to the facial action coding system (FACS; \citealt{eckman2002facs}).
\end{itemize}
In our experiments, we used frontal images and seven emotions from each of these datasets. An edge map illustration of landmarks for different emotions is shown in Fig.~\ref{em_illustration}.

At this point, it is worth recalling that we are learning a function-valued function,  $h: \mX \mapsto (\Theta \mapsto \mX)$ using a vRKHS as our hypothesis class (see Section~\ref{section:problem_setting}).
In the following we detail the choices made concerning the representation of the landmarks in $\mX$, that of the emotions in $\Theta$, and in the kernel design $k_{\mX}, k_{\Theta}$ and $\b A$.

\tb{Landmark representation, pre-processing:} We applied the following pre-processing steps to get the landmark representations which form the input of the algorithms. To extract $68$ landmark points for all the facial images, we used the standard \texttt{dlib} library. The estimator is based on \texttt{dlib}'s implementation of \cite{kazemi2014one}, trained on the iBUG 300-W face landmark dataset. Each landmark is represented by its 2D location. The alignment of the faces was carried out by the Python library \texttt{imutils}. The method ensures that faces across all identities and emotions are vertical, centered and of similar sizes. In essence, this is implemented through an affine transformation computed after drawing a line segment between the estimated eye centers. Each image was resized to the size $128 \times 128$. The landmark points computed in the step above were transformed through the same affine transformation. These two preprocessing steps gave rise to the aligned, scaled and vectorized landmarks $\b x \in \R^{136=2\times 68}$.

\begin{figure}
	\centering
	\includegraphics[scale=0.55]{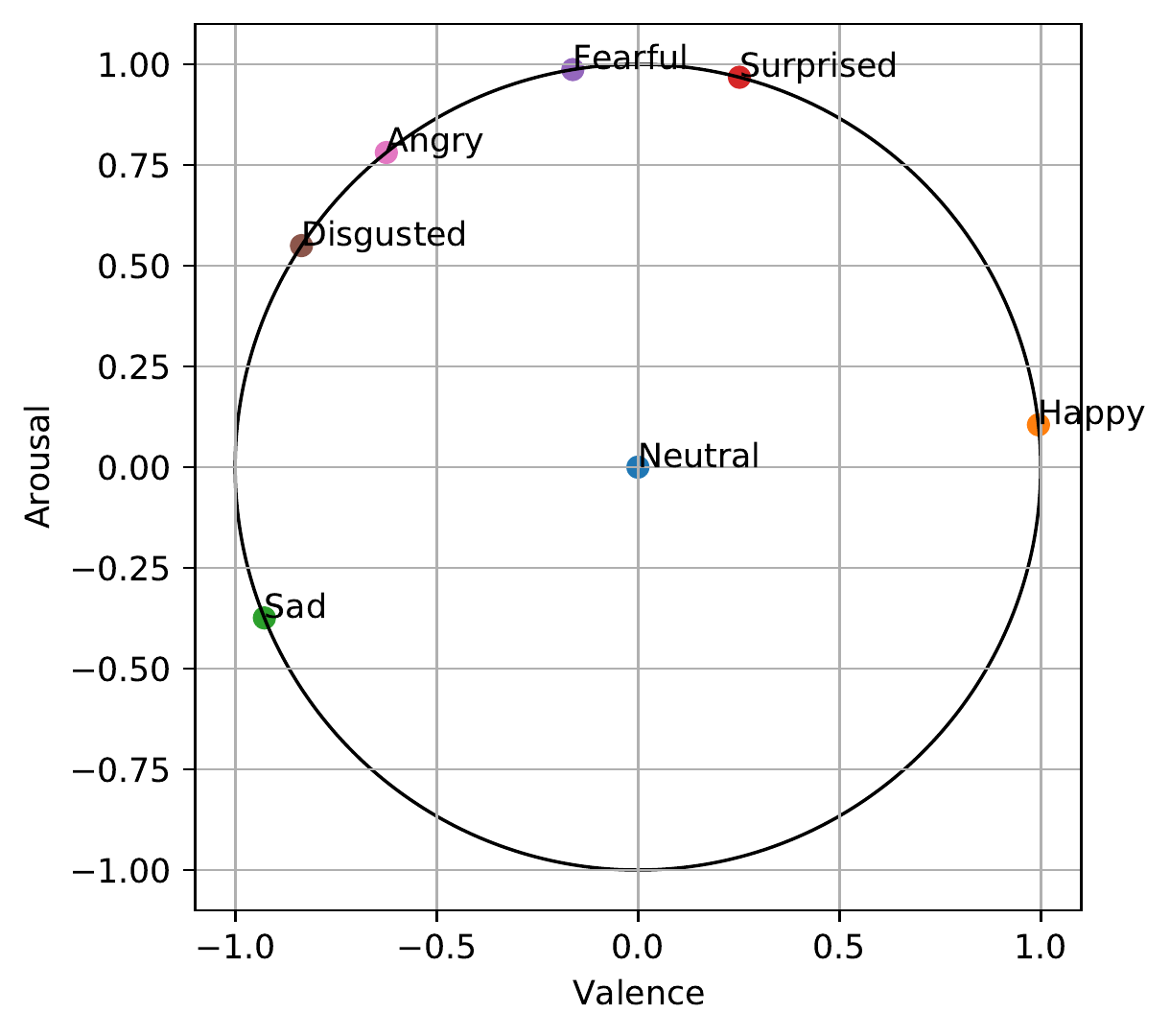}
	\caption{Extracted $\ell_2$-normalized valence-arousal centroids for each emotion from the manually annotated train set of the AffectNet database.}
	\label{va-space2}
\end{figure}

\tb{Emotion representation:} We represented emotion labels as points in the 2D valence-arousal space (VA, \citealt{russell1980circumplex}). Particularly, we used a  manually annotated part of the large-scale AffectNet database \citep{mollahosseini2017affectnet}. For all samples of a particular emotion in the AffectNet data, we computed the centroid (data mean) of the valence and arousal values. The resulting $\ell_2$-normalized 2D vectors constituted our emotion representation as depicted in Fig.~\ref{va-space2}. The normalization is akin to assuming that the modeled emotions are of the same intensity. In our experiments, the emotion `neutral' was represented by the origin. Such an emotion embedding allowed us to take into account prior knowledge about the angular proximity of emotions in the VA space, while keeping the representation simple and interpretable for post-hoc manipulations.

\tb{Kernel design:}
We took the kernels $k_{\mX}$, $k_{\Theta}$  to be Gaussian on the landmark representation space and the emotion representation space, with respective bandwidth $\gamma_{\mX}$ and $\gamma_{\Theta}$. $\b A$ was assumed to be $\b I_d$ unless specified otherwise.

\subsection{Quantitative Performance Assessment}
\label{expt:exp1}
In this section we provide a quantitative assessment of the proposed vITL approach.

\tb{Performance measures:} We applied two metrics to quantify the performance of the compared systems, namely the test mean squared error (MSE) and emotion classification accuracy. The classification accuracy can be thought of as an indirect evaluation. To compute this measure, for each dataset we trained  a ResNet-18 classifier to recognize emotions from ground-truth landmark edge maps (as depicted in Fig.~\ref{em_illustration}). The trained network was then used to compute classification accuracy over the predictions at test time. To rigorously evaluate outputs for each split of the data, we used a classifier trained on RaFD to evaluate KDEF predictions and vice-versa; this also allowed us to make the problem more challenging. The ResNet-18 network was appropriately modified to take grayscale images as input. During training, we used random horizontal flipping and cropping between 90-100\% of the original image size to augment the data. All the images were finally resized to $224 \times 224$ and fed to the network. The network was trained from scratch using the stochastic gradient descent optimizer with learning rate and momentum set to $0.001$ and $0.9$, respectively. The training was carried out for $10$ epochs with a batch size of $16$.

We report the mean and standard deviation of the aforementioned metrics over ten 90\%-10\% train-test splits of the data. The test set for each split is constructed by removing $10\%$ of the identities from the data. For each split, the best $\gamma_{\mX}, \gamma_{\Theta}$ and $\lambda$ values were determined by $6$-fold and $10$-fold cross-validation on KDEF and RaFD, respectively.

\textbf{Baseline:} We used the popular StarGAN \citep{choi2018stargan} system as our baseline. Other GAN-based studies use additional information and are not directly comparable to our setting. For fair comparison, the generator $G$ and discriminator $D$ were modified to be fully-connected networks that take vectorized landmarks as input. In particular, $G$ was an encoder-decoder architecture where the target emotion, represented as a 2D emotion encoding as for our case, was appended at the bottleneck layer. It contained approximately one million parameters, which was chosen to be comparable with the number of coefficients in vITL ($839,664 = 126 \times 7 \times 7 \times 136$ for KDEF). ReLU activation function was used in all layers except before bottleneck in $G$ and before penultimate layers of both $G$ and $D$. We used their default parameter values in the code.\footnote{The code is available at \url{https://github.com/yunjey/stargan}.}  Experiments over each split of KDEF and RaFD were run for 50K and 25K iterations, respectively.

\tb{MSE results:} The test MSE for the compared systems is summarized in Table~\ref{tab:test_risk}. As the table shows, the vITL technique outperforms StarGAN on both datasets. One can observe low reconstruction cost for vITL in both the single and the joint emotional input case. Interestingly, a performance gain is obtained with vITL joint on the RaFD data in MSE sense. We hypothesize that this is due to the joint model benefiting from input landmarks for other emotions in the small data regime (only $67$ samples per emotion for RaFD).
Despite our best efforts, we found it quite difficult to train StarGAN reliably and the diversity of its outputs was low.

\begin{figure}
    \begin{subfigure}[b]{0.45\textwidth}
  \includegraphics[scale=0.25]{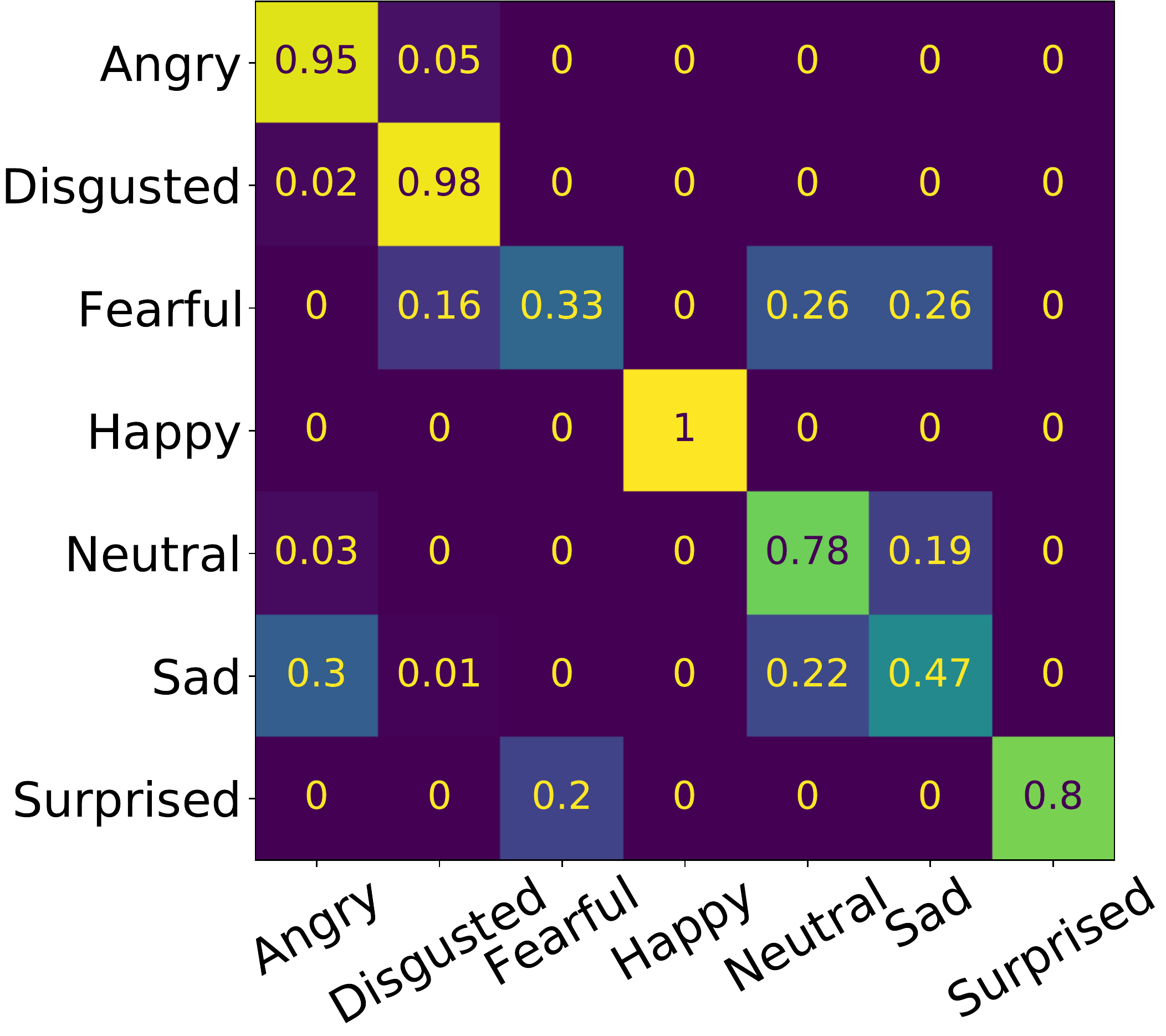}
  \caption{KDEF}
    \end{subfigure}~
    \begin{subfigure}[b]{0.45\textwidth}
  \includegraphics[scale=0.25]{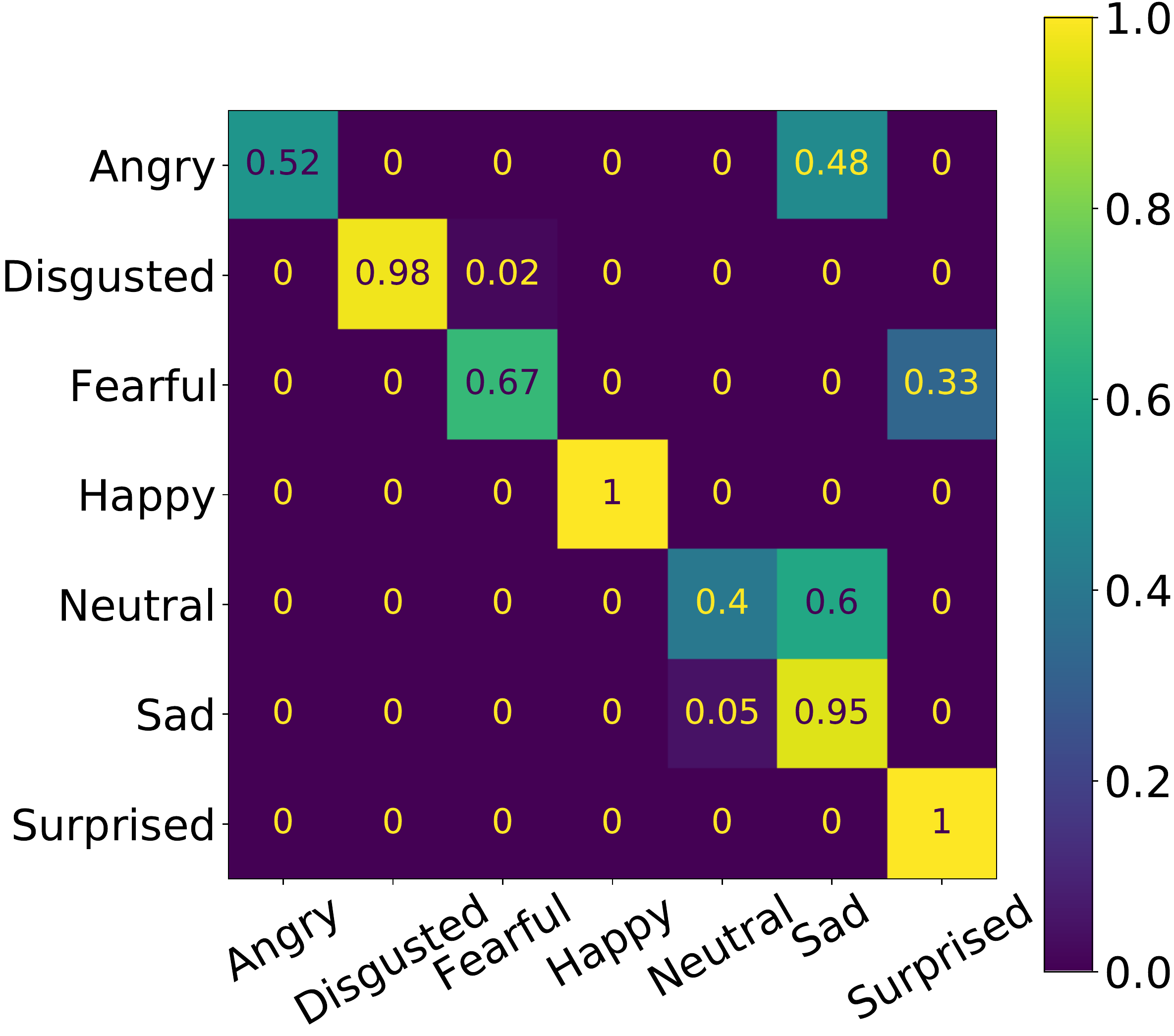}
  \caption{RaFD}
    \end{subfigure}
	\caption{Confusion matrices for classification accuracy of vITL Joint model. Left: dataset KDEF. Right: dataset RaFD. The $y$ axis represents the true labels, the $x$ axis stands for the predicted labels. More diagonal is better.}
	\label{fig:confusion_matrix}
\end{figure}

\tb{Classification results:} The emotion classification accuracies are available in  Table~\ref{tab:emo-classification}. The classification results clearly demonstrate the improved performance and the higher quality of the generated emotion of vITL over StarGAN; the latter also produces predictions with visible face distortions as it is illustrated in Section~\ref{expt:qualitative}.
To provide further insight into the classification performance we also show the confusion matrices for the joint vITL model on a particular split of KDEF and RaFD datasets in Fig.~\ref{fig:confusion_matrix}. For both the datasets, the  classes `happy' and `surprised' are easiest to detect. Some confusions arise between the classes `neutral' vs `sad' and `fearful' vs `surprised'. Such mistakes are expected when only using landmark locations for recognizing emotions.

\begin{table}
	\centering
	\begin{tabular}{l c c}
		\toprule
		Methods& \multicolumn{1}{c}{KDEF frontal} & \multicolumn{1}{c}{RaFD frontal} \\
		\midrule

		vITL: $\theta_0=$ neutral & $0.010 \pm  0.001$ & $0.009 \pm  0.004$ \\
		vITL: $\theta_0=$ fearful & $0.010 \pm  0.001$  & $0.010 \pm  0.005$  \\
		vITL: $\theta_0=$ angry & $0.012 \pm  0.002$ & $0.010 \pm  0.005$ \\
		vITL: $\theta_0=$ disgusted & $0.012 \pm  0.001$ & $0.010 \pm  0.004$  \\
		vITL: $\theta_0=$ happy & $0.011 \pm  0.001$ & $0.010 \pm  0.004$  \\
		vITL: $\theta_0=$ sad &$0.011 \pm  0.001$  & $0.009 \pm  0.004$\\
		vITL: $\theta_0=$ surprised &$0.010 \pm  0.001$  & $0.011 \pm  0.006$ \\
		\midrule
		vITL: Joint & $0.011 \pm 0.001$ & $0.007 \pm 0.001$\\
		\midrule
		StarGAN &$0.029 \pm 0.003$& $0.024 \pm 0.007$ \\
		\bottomrule
	\end{tabular}
	\caption{MSE error (mean $\pm$ std) on test data for the vITL single (top), the vITL joint and the StarGAN system (bottom). Lower is better.}
	\label{tab:test_risk}
\end{table}

\begin{table}
	\centering
	\begin{tabular}{l c c}
		\toprule
		Methods& \multicolumn{1}{c}{KDEF frontal} & \multicolumn{1}{c}{RaFD frontal} \\
		\midrule
		vITL: $\theta_0=$ neutral & $76.12 \pm  4.57$  & $79.76 \pm  7.88$ \\
		vITL: $\theta_0=$ fearful & $76.22 \pm  4.91$  & $78.81 \pm  8.36$  \\
		vITL: $\theta_0=$ angry &$74.49 \pm  2.31$  & $78.10 \pm  7.51$ \\
		vITL: $\theta_0=$ disgusted &$74.18 \pm  4.22$  & $78.33 \pm  4.12$  \\
		vITL: $\theta_0=$ happy &$73.57 \pm  2.74$  & $80.48 \pm  5.70$  \\
		vITL: $\theta_0=$ sad &$75.82 \pm  4.11$  & $77.62 \pm  5.17$\\
		vITL: $\theta_0=$ surprised &$74.69 \pm  2.25$  & $80.71 \pm  5.99$ \\
		\midrule
		vITL: Joint & $74.81 \pm 3.10$  & $77.11 \pm 3.97$ \\
		\midrule
		StarGAN & $70.69 \pm 8.46$ & $65.88 \pm 8.92$ \\
		\bottomrule
	\end{tabular}
	\caption{Emotion classification accuracy (mean $\pm$ std) for the vITL single (top), the vITL joint (middle) and the StarGAN system (bottom). Higher is better.}
	\label{tab:emo-classification}
\end{table}

\subsection{Analysis of Additional Properties of vITL} \label{expt:exp2}
This section is dedicated to the effect of the choice of $\b A$ (in kernel $G$) and to the robustness of vITL w.r.t.\ partial observation.

\begin{figure}
	\centering
	\includegraphics[scale=0.5]{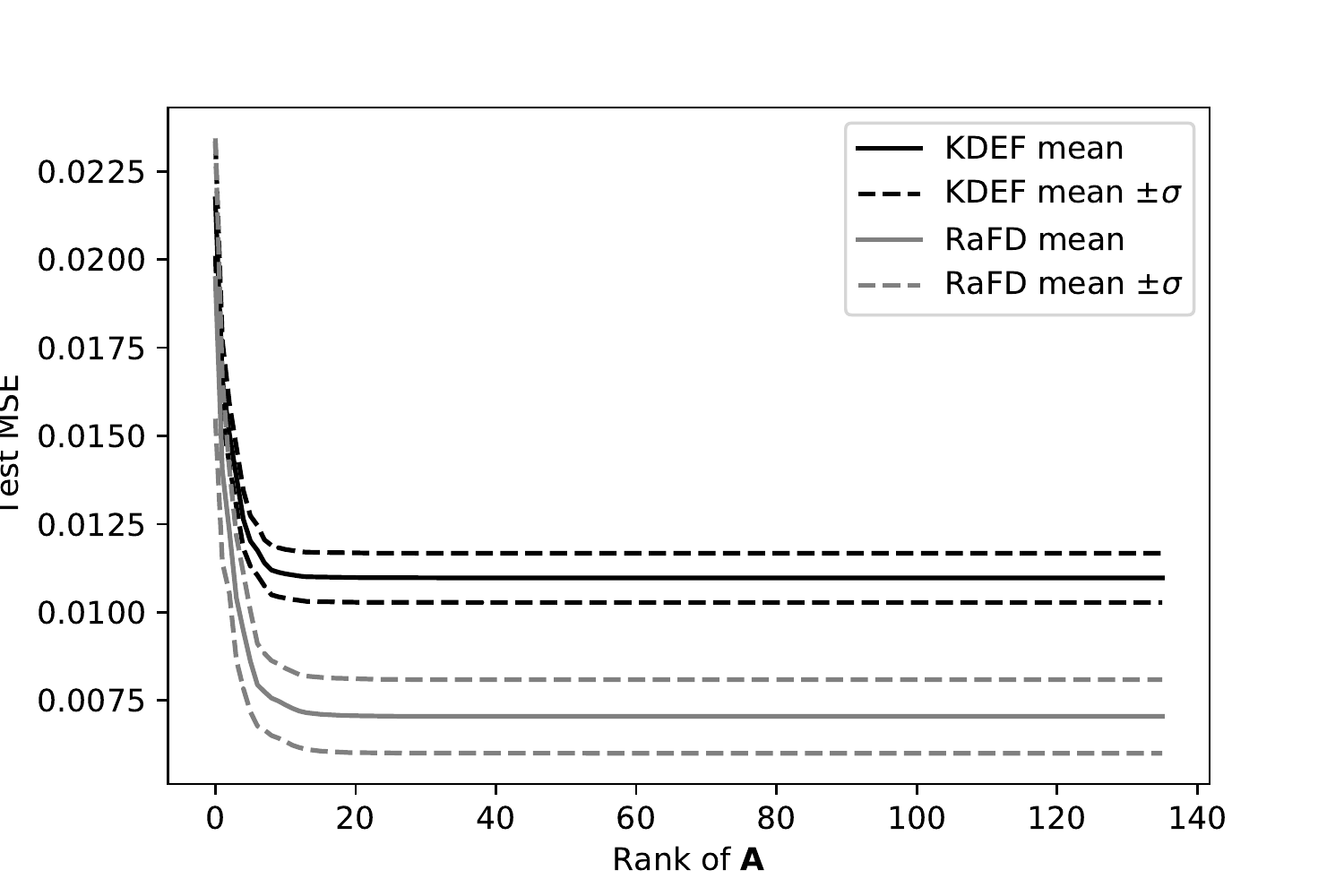}
	\caption{Test MSE (mean $\pm$ std) as a function of the rank of the matrix $\b{A}$. Smaller MSE is better.}
	\label{fig:matrix}
\end{figure}

\tb{Influence of $\b A$ in the matrix-valued kernel $G$:} Here, we illustrate the effect of matrix $\b A$ (see \eqref{eq:G,K:def}) on the vITL estimator and show that a good choice of $\b A$ can lead to lower dimensional models, while preserving the quality of the prediction. The choice of $\b A$ is built on the knowledge that the empirical covariance matrices of the output training data contains structural information that can be exploited with vRKHS \citep{kadri2013generalized}. In order to investigate this possibility, we performed the singular value decomposition of $\b{Y}^\top \b{Y}$ which gives the eigenvectors collected in matrix $\b{V}\in \R^{d\times d}$. For a fixed rank $r \le d$, define $\b{J}_r = \text{diag}(\underbrace{1, \cdots, 1}_{r}, \underbrace{0, \cdots, 0}_{d-r})$, set $\b{A} = \b{V} \, \b{J}_r \, \b{V}^\top$ and train a vITL system with the resulting $\b A$. While in this case $\bf A$ is no more invertible, each coefficient $\hat{c}_{i,j}$ from Lemma~\ref{lemma:double-repr} belongs to the $r$-dimensional subspace of $\R^d$ generated by the eigenvectors associated to the $r$ largest eigenvalues of $\b{Y}^\top \b{Y}$. This makes a reparameterization possible and leads to a decrease in the size of the model, going from $t \times m \times d$ parameters to $t \times m \times r$. We report in Fig.~\ref{fig:matrix} the resulting test MSE performance (mean $\pm$ standard deviation) obtained from $10$ different splits, and empirically observe that $r=20$ suffices to preserve the optimal performances of the model.

\begin{figure}
	\centering
	\includegraphics[scale=0.5]{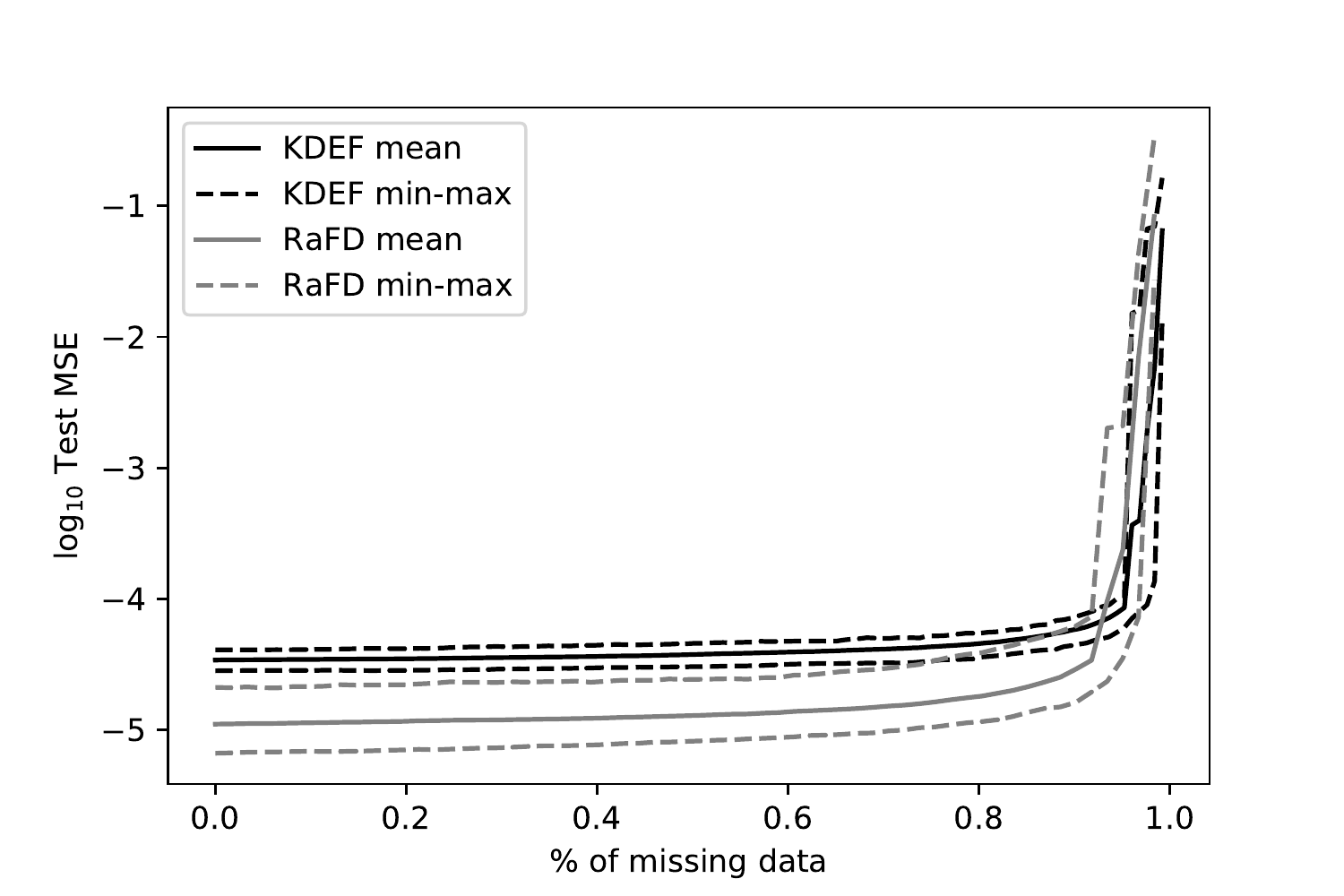}
	\caption{Logarithm of the test MSE (min-mean-max) as a function of the percentage of missing data. Solid line: mean; dashed line: min-max. Smaller MSE is better.}
	\label{fig:partial}
\end{figure}

\tb{Learning under a partial observation regime:} To assess the robustness of vITL w.r.t.\ missing data, we considered a random mask $(\eta_{i,j})_{i \in [n], j \in [m]} \in \{0,1\}^{n \times m}$; a sample  $z_{i}(\theta_{i,j})$ was used for learning only when $\eta_{i,j}=1$. Thus, the percentage of missing data was $p:=\frac{1}{nm} \sum_{i, j \in [n]\times [m]} \eta_{i,j}$. The experiment was repeated for $10$ splits of the dataset, and on each split we averaged the results using $4$ different random masks $(\eta_{i,j})_{i \in [n], j \in [m]}$. The resulting test MSE of the predictor as a function of $p$ is summarized in Fig.~\ref{fig:partial}. As it can be seen, the vITL approach is quite stable in the presence of missing data on both datasets.

\subsection{Qualitative Analysis} \label{expt:qualitative}
In this section we show example outputs produced by vITL in the context of discrete and continuous emotion generation. While the former is the classical task of synthesis given input landmarks and target emotion label, the latter serves to demonstrate a key benefit of our approach, which is the ability to synthesize meaningful outputs while continuously traversing the emotion embedding space.

\begin{figure}
    \centering
    \includegraphics[scale=0.43]{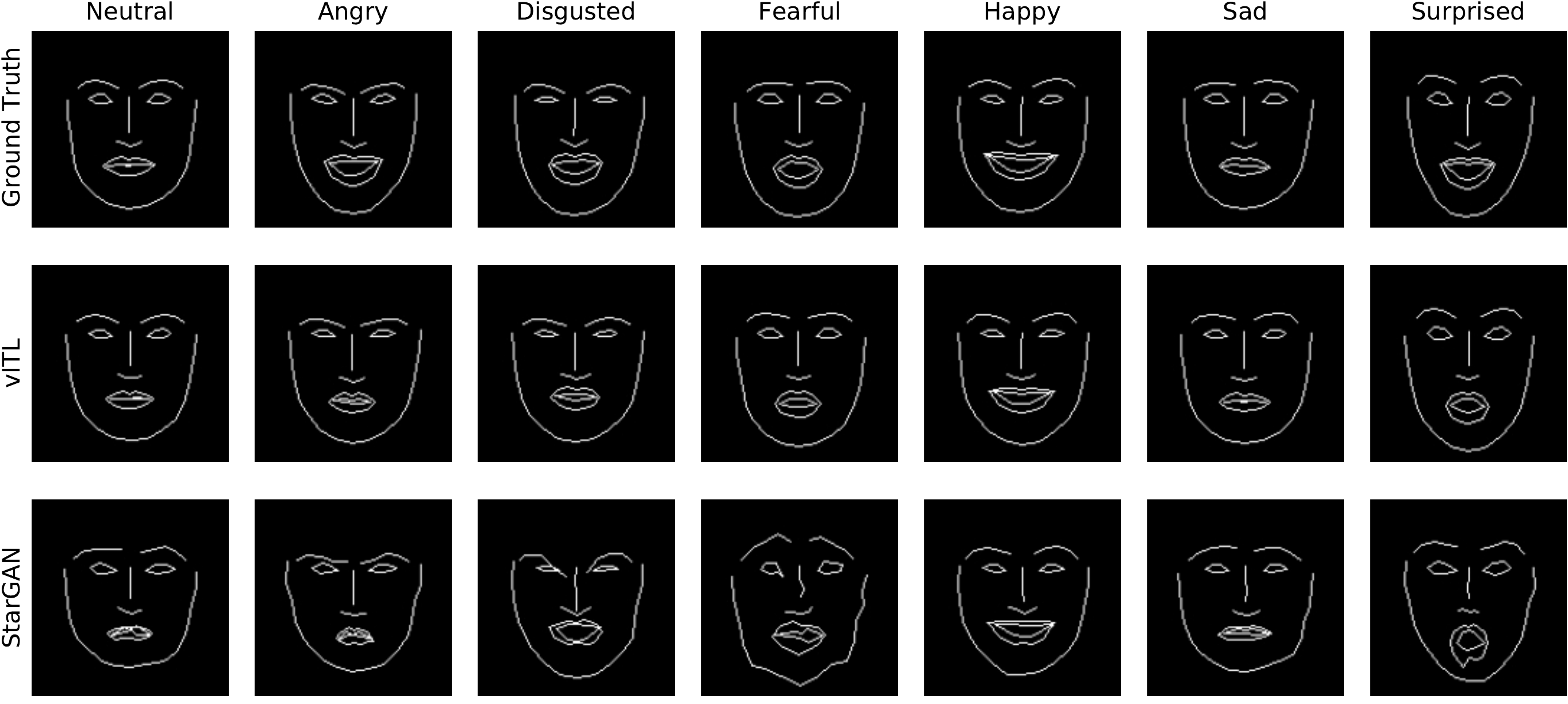}
    \caption{Discrete expression synthesis results on the KDEF dataset with ground-truth neutral landmarks as input.}
    \label{fig:kdef_discrete}
\end{figure}

\begin{figure}
    \centering
    \includegraphics[scale=0.43]{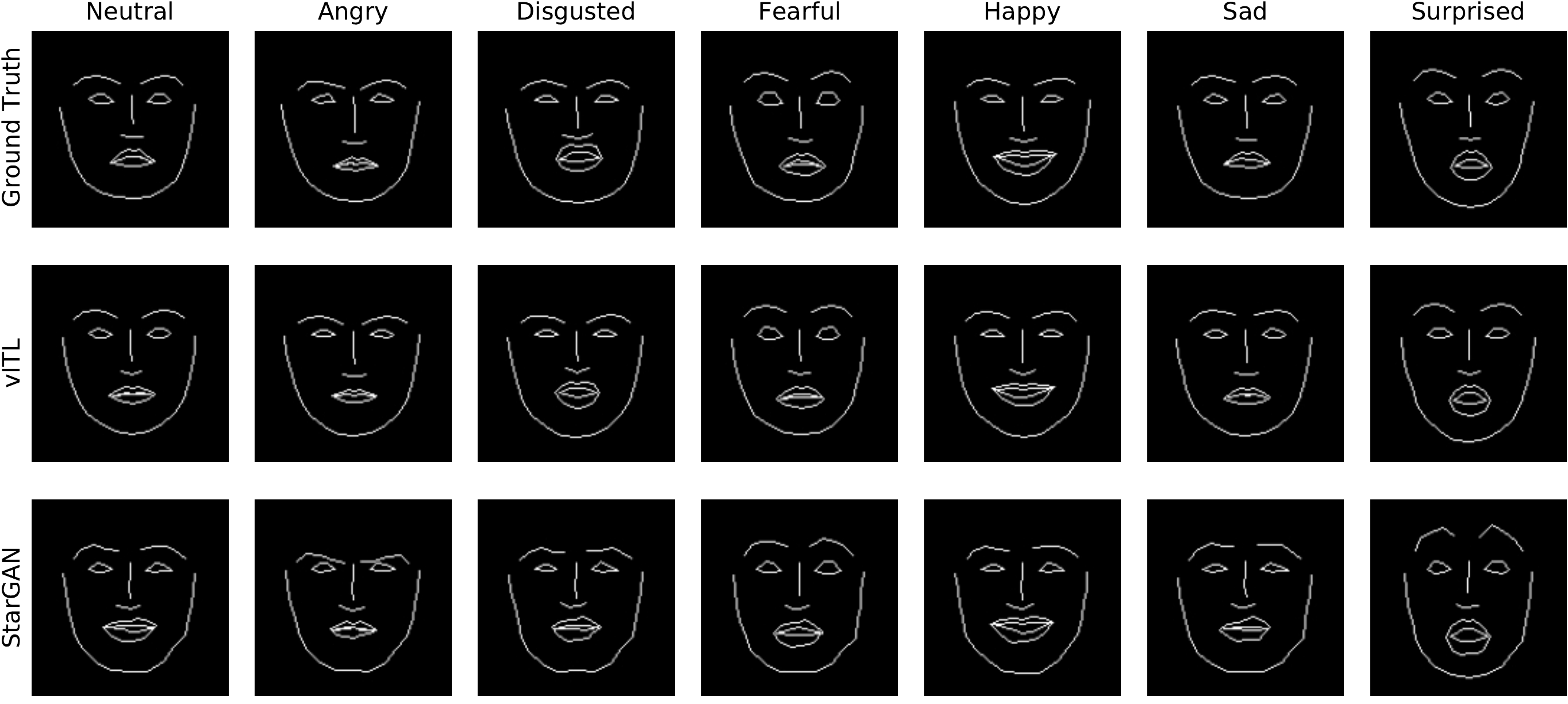}
    \caption{Discrete expression synthesis results on the RaFD dataset with ground-truth neutral landmarks as input.}
    \label{fig:rafd_discrete}
\end{figure}

\noindent\tb{Discrete emotion generation:}
In Fig. \ref{fig:kdef_discrete} and \ref{fig:rafd_discrete} we show qualitative results for generating landmarks using discrete emotion labels present in the datasets. For vITL, not only are the emotions recognizable, but landmarks on the face boundary are reasonably well synthesized and other parts of the face visibly less distorted when compared to StarGAN. The identity in terms of the face shape is also better preserved.

\begin{figure}
    \centering
    \includegraphics[scale=0.5]{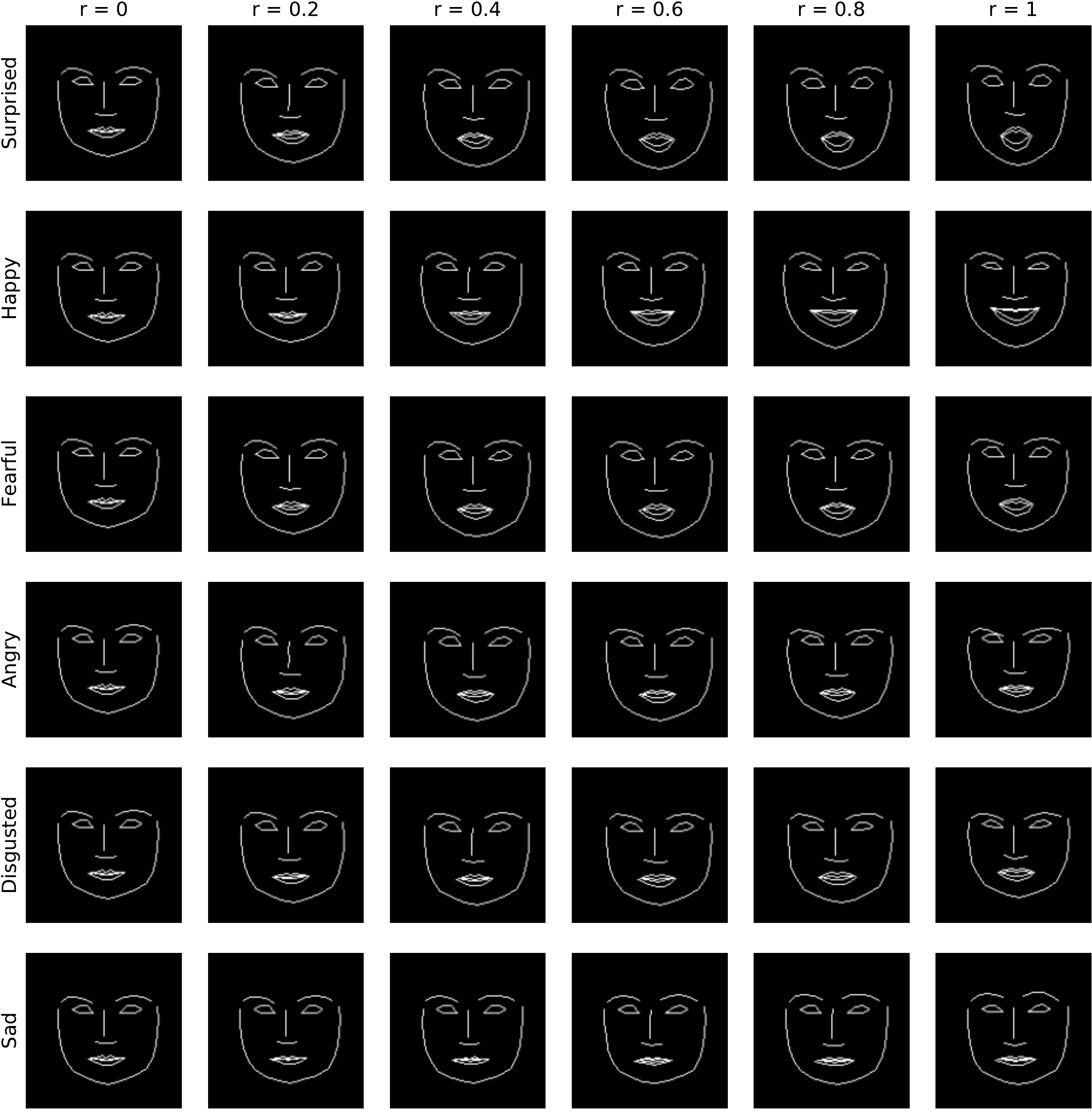}
    \caption{Continuous expression synthesis results with vITL on the KDEF dataset, with ground-truth neutral landmarks. The generation is starting from neutral and proceeds in the radial direction towards an emotion with increasing radii $r$.}
    \label{fig:kdef_continuous}
\end{figure}

\begin{figure}
    \centering
    \includegraphics[scale=0.5]{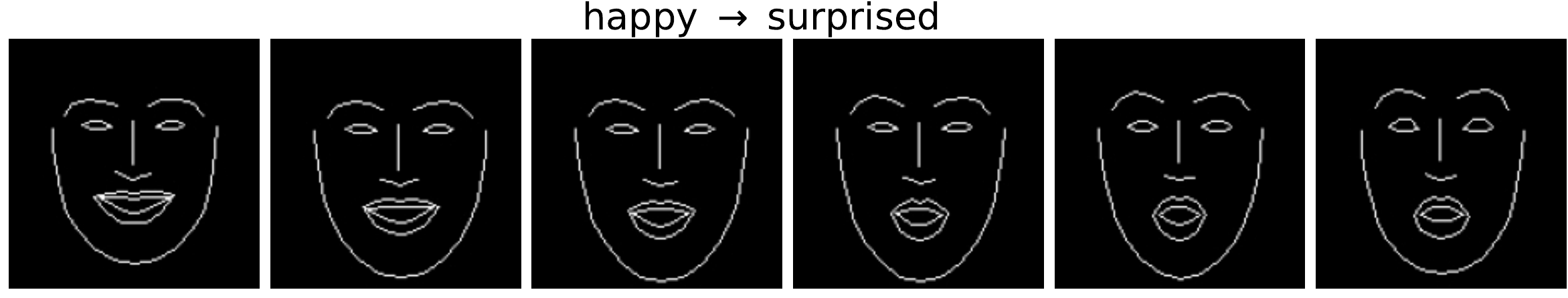}
    \caption{Continuous expression synthesis with vITL technique on the RaFD dataset, with ground-truth neutral landmarks. The generation is starting from `happy' and proceeds by changing angular position towards `surprised'. For a more fine-grained video illustration traversing from `happy' to `sad' along the circle, see the demo on \href{https://www.github.com/allambert/torch_itl}{GitHub}.}
    \label{fig:rafd_circ}
\end{figure}

\noindent \tb{Continuous emotion generation:}
Starting from neutral emotion, continuous generation in the radial direction is illustrated in Fig.~\ref{fig:kdef_continuous}. The landmarks vary smoothly and conform to the expected intensity variation in each emotion on increasing the radius of the vector in VA space. We also show in Fig. \ref{fig:rafd_circ} the capability to generate intermediate emotions by changing the angular position, in this case from `happy' to `surprised'. For a more fine-grained video illustration traversing from `happy' to `sad' along the circle, see the \href{https://www.github.com/allambert/torch_itl}{GitHub} repository.

These experiments and qualitative results demonstrate the efficiency of the vITL approach in emotion transfer.

\section{Conclusion}
In this paper we introduced a novel approach to style transfer based on function-valued regression, and exemplified it on the problem of emotion transfer. The proposed  vector-valued infinite task learning (vITL) framework relies on operator-valued kernels. vITL (i) is  capable of encoding and controlling continuous style spaces, (ii) benefit from a representer theorem for efficient computation, and (iii) facilitates regularity control via the choice of the underlying kernels. The framework can be extended in several directions. Other losses \citep{sangnier16joint,laforgue2020duality} can be leveraged to produce outlier-robust or sparse models.  
Instead of being chosen prior to learning, the input kernel could be learned using deep architectures \citep{mehrkanoon2018,liu20learning} opening the door to a wide range of applications.

\section{Proofs}
This section contains the proofs of our auxiliary lemmas. \begin{proof}{(Lemma~\ref{lemma:double-repr})}
For all $g \in \mH_G$, let $K_x g$ denote the function defined by $(K_x g)(t) = K(t,x)g$ $\forall t \in \mX$. Similarly, for all $c \in \mX$, $G_{\theta}c$ stands for the function $t \mapsto G(t,\theta)c$ where $t\in \Theta$. Let us take the finite-dimensional subspace 
\begin{align*}
 E = \text{span}\left(K_{x_i} G_{\theta_{ij}} c \, : \, i \in [t] , j \in [m], c \in \R^d \right).   
\end{align*}
The space $\mH_K$ can be decomposed as $E$ and its orthogonal complement: $E \oplus E^\perp = \mH_K$. 
The existence of $\hat{h}$ follows from the coercivity of $\mR_{\lambda}$ (i.e.\ $\mR_{\lambda}(h) \to + \infty$ as $\left\|h\right\|_{\mH_K} \to +\infty$) 
which is the consequence of the quadratic regularizer and the lower boundedness of $\ell$.
Uniqueness comes from the strong convexity of the objective.
Let us decompose $\hat{h} = \hat{h}_E + \hat{h}_{E^\perp}$, and take any $c \in \R^d$. Then $\forall (i,j) \in [t] \times [m]$, 
\begin{align*}
    \left \langle \hat{h}_{E^\perp}(x_i)(\theta_{ij}), c \right \rangle_{\reals^d} \stackrel{(a)}{=} 
    \left \langle \hat{h}_{E^\perp}(x_i) ,  G_{\theta_{ij}} c \right \rangle_{\mH_G} \stackrel{(b)}{=}  \big< \hat{h}_{E^\perp} , \underbrace{K_{x_i} G_{\theta_{ij}} c}_{\in E}  \big>_{\mH_K} \stackrel{(c)}{=} 0.
\end{align*}
(a) follows from the reproducing property in $\mH_G$, 
(b) is a consequence of the reproducing property in $\mH_K$, and (c) comes from the decomposition $E \oplus E^\perp = \mH_K$. This means that $\hat{h}_{E^\top}(x_i)(\theta_{ij}) = 0$ $\forall (i,j) \in [t] \times [m]$, and hence $\riskemp(\hat{h}) = \riskemp(\hat{h}_E)$.
Since $ \lambda \big\| \hat{h} \big\|_{\mH_K}^2 = \lambda \left( \big\| \hat{h}_E \big\|_{\mH_K}^2 + \big\| \hat{h}_{E^\perp} \big\|_{\mH_K}^2 \right) \ge \lambda \big\| \hat{h}_E \big\|_{\mH_K}^2$ 
we conclude that $\hat{h}_{E^\top} = 0$ and get that there exist coefficients $\hat{c}_{i,j} \in \R^d$ such that $\hat{h} = \sum_{i \in [t]} \sum_{j \in [m]} K_{x_i} G_{\theta_{i,j}} \hat{c}_{i,j}$. This evaluates for all $(x,\theta) \in \mX \times \Theta$ to 
\begin{align*}
\hat{h}(x)(\theta) = \sum_{i=1}^t \sum_{j=1}^m k_{\mX}(x,x_i) k_{\Theta}(\theta,\theta_{i,j}) \mathbf{A} \hat{c}_{ij}
\end{align*}
as claimed in \eqref{eq:hx-theta}.
\end{proof}

\begin{proof}{(Lemma~\ref{thm:krr})}
Applying Lemma~\ref{lemma:double-repr}, problem \eqref{emp-ivtl} writes as \begin{equation*}
  \min_{\b{C} \in \R^{(tm)\times d}} \frac{1}{2tm} \left\| \b{KCA} - \b{Y}\right\|_{\text{F}}^2 + \frac{\lambda}{2} \Tr \left( \b{KCAC^\top} \right),
\end{equation*}
where $\left\|\cdot\right\|_{\text{F}}$ denotes the Frobenius norm.
By setting the gradient of this convex functional to zero, and using the symmetry of $\b{K}$ and $\b{A}$, one gets
\begin{align*}
  \frac{1}{tm} \mathbf{K}(\mathbf{KCA} - \mathbf{Y})\mathbf{A} + \lambda \mathbf{KCA} = \b 0
\end{align*}
which implies \eqref{eq:sylvester} by the invertibility of $\b{K}$ and $\b A$.
\end{proof}

\paragraph{Acknowledgements} A.L.\ and S.P.\ were funded by the research chair \href{https://datascienceandai.wp.imt.fr/en/home-2/}{Data Science \& Artificial Intelligence for Digitalized Industry and Services} at T{\'e}l{\'e}com Paris. ZSz benefited from the support of the Europlace Institute of Finance and that of the \href{http://www.cmap.polytechnique.fr/~stresstest/}{Chair Stress Test}, RISK Management and Financial Steering, led by the French École Polytechnique and its Foundation and sponsored by BNP Paribas.

\bibliographystyle{plainnat}
\bibliography{egbib}

\end{document}